\documentclass[letterpaper, 10 pt, conference]{ieeeconf}
\IEEEoverridecommandlockouts
\let\labelindent\relax
\usepackage{enumitem}
\usepackage{balance}
\usepackage[font={footnotesize}]{caption}
\usepackage{subcaption}
\usepackage{array}
\usepackage{textcomp}
\usepackage{mathtools, nccmath}
\usepackage{graphicx}
\usepackage{amsfonts}
\usepackage{amsmath}
\usepackage{amssymb}
\usepackage{algorithm}
\usepackage{algorithmicx}
\usepackage{tikz}
\usepackage{arydshln}
\usepackage{multirow}
\usepackage{bm}
\usepackage{epstopdf}
\usepackage{siunitx}
\usepackage{xcolor}
\usepackage{tabularx}
\usepackage{cite}
\usepackage{listings}
\usepackage{algpseudocode}
\usepackage{bbold}
\usepackage{hyperref}
\usepackage{xcolor}
\usepackage{booktabs}       
\usepackage{tabularx}

\makeatletter
\let\NAT@parse\undefined
\makeatother
\newcommand{\subfig}[1]{\textit{\footnotesize{#1}}}

\newcommand{\figref}[1]{Fig.~\ref{#1}}

\newcommand{\RobiButler}[0]{\textit{Robi Butler}}

\newcommand{\GestureOnly}[0]{\textit{Gesture-only}}
\newcommand{\VoiceOnly}[0]{\textit{Voice-only}}
\newcommand{\GesturePlusVoice}[0]{\textit{Gesture+Voice}}

\newcommand{\pval}[1]{$p#1$}

\newcommand{\significantI}[0]{$^{\star}$}

\newcommand{\RQI}[0]{\textit{\textbf{RQ1}}}
\newcommand{\RQII}[0]{\textit{\textbf{RQ2}}}

\renewcommand{\quote}[1]{``\textit{#1}''}

\newif\ifcomment
\commenttrue 

\ifcomment

\newcommand{\todo}[1]{\textcolor{red}{TODO: #1}}

\else

\newcommand{\todo}[1]{}

\fi

\definecolor{codegreen}{rgb}{0,0.6,0}
\definecolor{codegray}{rgb}{0.5,0.5,0.5}
\definecolor{codepurple}{rgb}{0.58,0,0.82}
\definecolor{backcolour}{rgb}{0.95,0.95,0.92}

\lstdefinestyle{mystyle}{
    backgroundcolor=\color{backcolour},   
    commentstyle=\color{codegreen},
    keywordstyle=\color{magenta},
    numberstyle=\tiny\color{codegray},
    stringstyle=\color{codepurple},
    basicstyle=\ttfamily\footnotesize,
    breakatwhitespace=false,         
    breaklines=true,                 
    captionpos=b,                    
    keepspaces=true,                 
    numbers=left,                    
    numbersep=5pt,                  
    showspaces=false,                
    showstringspaces=false,
    showtabs=false,                  
    tabsize=2
}

\title{\LARGE \bf \RobiButler{}: Multimodal Remote Interaction with a \\Household Robot Assistant
}

\author{
Anxing~Xiao, 
Nuwan~Janaka,
Tianrun~Hu,
Anshul~Gupta, 
Kaixin~Li,
Cunjun~Yu,
David~Hsu
\thanks{All authors are with the School of Computing, National University of Singapore, Singapore.
Correspond to \tt\small anxingx@comp.nus.edu.sg.
}
\thanks{Nuwan~Janaka, Tianrun~Hu, and David~Hsu are also with the Smart Systems Institute, National University~of Singapore, Singapore.}
}

\begin{document}

\maketitle
\begin{abstract}
Imagine a future when we can Zoom-call a robot to manage household chores remotely. 
This work takes one step in this direction. \textit{Robi Butler} is a new household robot assistant that enables seamless multimodal remote interaction. 
It allows the human user to monitor its environment from a first-person view, issue voice or text commands, and specify target objects through hand-pointing gestures. 
At its core, a high-level behavior module, powered by Large Language Models (LLMs), interprets multimodal instructions to generate multistep action plans. 
Each plan consists of open-vocabulary primitives supported by vision-language models, enabling the robot to process both textual and gestural inputs. 
Zoom provides a convenient interface to implement remote interactions between the human and the robot.
The integration of these components allows Robi Butler to ground remote multimodal instructions in real-world home environments in a zero-shot manner. 
We evaluated the system on various household tasks, demonstrating its ability to execute complex user commands with multimodal inputs. 
We also conducted a user study to examine how multimodal interaction influences user experiences in remote human-robot interaction. 
These results suggest that with the advances in robot foundation models, we are moving closer to the reality of remote household robot assistants.
Link: \url{https://robibutler.github.io/}.

\end{abstract}

\IEEEpeerreviewmaketitle


\section{Introduction}
\label{sec:introduction}

\begin{figure}[t]
  \centering
  \includegraphics[width=0.95\linewidth]{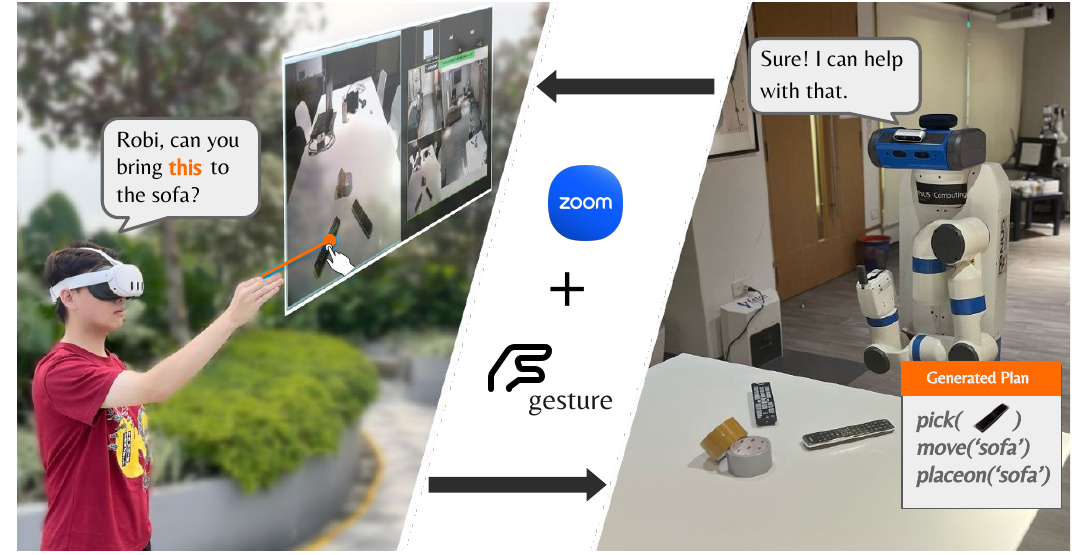}
  \caption{The \RobiButler{}  system enables the user to Zoom-call the butler robot remotely at home and interact with it naturally through both the \textit{language} and \textit{hand gestures}. 
  }
  \label{fig:main}
\vspace{-0.6cm}
\end{figure}

Imagine a future where distance no longer constrains our ability to manage household tasks. Picture a robot assistant capable of remotely interpreting spoken commands and gestures to check your refrigerator or reheat a meal before you get home. Such a robotic system would fundamentally change the way we interact with our homes, bringing a new level of convenience and efficiency to daily life.
In this work, we propose \RobiButler{}, a multimodal interaction system that enables seamless communication between remote users and household robots to execute various household tasks. 
\RobiButler{} allows users to leverage both natural language and gestures to command the robot to perform tasks remotely, see~\figref{fig:main}. Remote users can point to the desired object in the MR device and instruct the robot to manipulate it, move toward it, or ask questions about it, just like a real butler.

The core issue behind building such a robot assistant is how to allow the robot to remotely \textit{receive}, \textit{understand}, and \textit{ground} the multimodal instructions into the executable actions in the home environment. 
To address this, we first design the communication interfaces consisting of a Zoom chat website and a gesture website for hand-pointing, which allows human users to send multimodal instructions using language and pointing remotely. To ground the received multimodal instructions in the home, the robot needs to have the ability to interpret and execute the open multimodal instructions in real-world environments. 
Inspired by the advanced capabilities of foundation models to achieve open vocabulary mobile manipulation in domestic environments \cite{brohan2023can,wu2023tidybot, yenamandrahomerobot,liu2024ok}, we aim to incorporate the LLM-based robots with the ability to make use of the language-related gestures. To allow the robot to ground both open language instruction and gesture selection, we first implement a mobile manipulation system that supports open vocabulary action primitives with pointing selection, driven by the recent advances in vision language models (VLMs).
Then, we introduce a high-level behavior module powered by large language models (LLMs), which organizes and aligns the received language and gesture instructions to generate the plan. 

Overall, the proposed system, \RobiButler{}, is a multimodal interactive system for robotic home assistants that enables bi-directional remote human-robot interaction based on the real home environment through text, voice, video, and gesture. We evaluated the performance of \RobiButler{} on real-world daily household tasks and studied the benefits of such multimodal interaction in terms of efficiency and user experience in the remote human-robot interaction.

\begin{figure*}[hptb]
  \centering
  \includegraphics[width=0.98\linewidth]{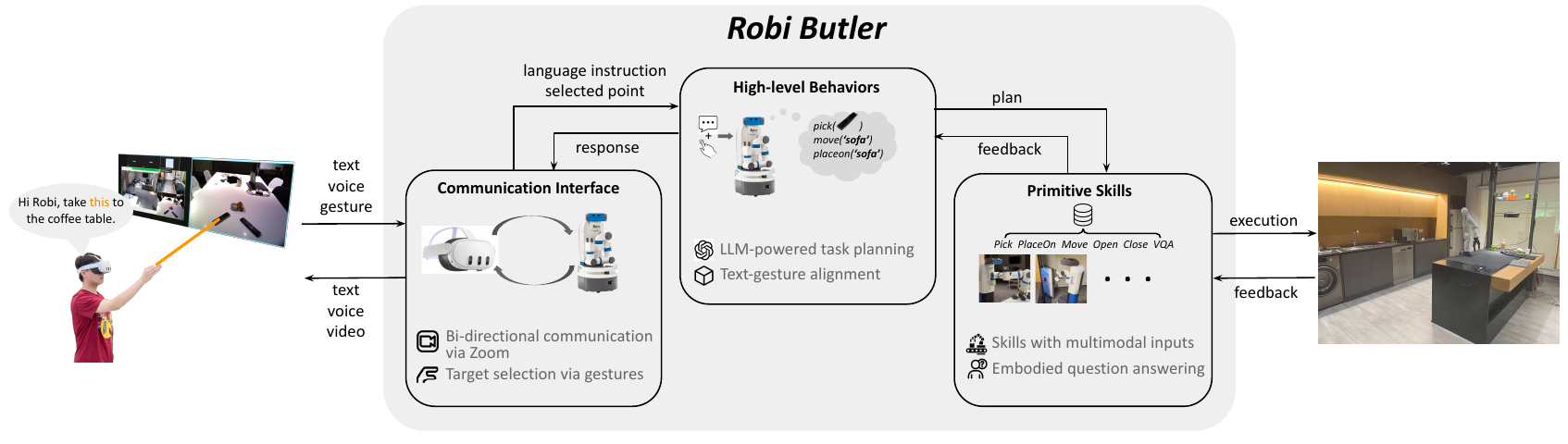}
  \caption{An overview of \emph{Robi Butler}. The system consists of three components: Communication Interface, High-level Behavior Module, and Primitive Skills. The Communication Interfaces transmit the inputs received from the remote user to the High-level Behavior Module, which composes the Primitive Skill to interact with the environment to fulfill the instructions or answer questions.
  }
  \label{fig:framework}
  \vspace{-0.6cm}
\end{figure*}

\section{Related Work}
\label{sec:relatedwork}

\subsection{Language and Gestures in Human-Robot Interaction}
Effective communication interfaces are essential for Human-Robot Interaction (HRI). Natural language instruction for robots has been widely explored in prior research, employing both traditional methods \cite{kollar2010toward, cantrell2010robust, tellex2011understanding, guadarrama2014open, misra2016tell, hatori2018interactively, paxton2019prospection, shridhar2020ingress, zhang2021invigorate} and large language models ~\cite{shridhar2022cliport, huang2022language, brohan2023can, huang2023inner, song2023llm, driess2023palm, rana2023sayplan, shi2024yell}. 
However, language can be ambiguous and imprecise. Humans typically use nonverbal interaction, such as pointing, to supplement their verbal instructions~\cite{diessel2020demonstratives}.
Previous work explores the use of tools such as laser pointers~\cite{Nguyen2008clickableworld} and point-and-click interfaces~\cite{kent2017comparison} to improve instruction delivery and further integrate both speech and relevant gestures together~\cite{matuszek2014learning, whitney2016interpreting, chen2021yourefit, weerakoon2022cosm2ic} to specify the command more precisely. 
However, these systems typically rely on predefined instruction templates or task-specific in-domain model training, which limits generalization to open-ended multi-modal language instruction.
Recent work uses LLMs to interpret gestures and commands~\cite{lin2023gesture}, but only handles short speech inputs and requires the user to be within the third-person camera view. Our system is built on top of a multimodal communication interface to construct a \textit{virtual clickable world} that allows the remote user to select the target by pointing while speaking, and the robot could interpret and execute the multimodal instructions in the home environment with a mobile manipulator.

\subsection{Household Robot Assistant}
Intelligent home robots with mobile manipulation capabilities can greatly expand functionality and integrate more seamlessly into daily routines. While past household mobile manipulation systems have been developed both in simulation~\cite{szot2021habitat, li2022igibson, srivastava2022behavior} and real-world settings~\cite{khatib1999mobile, reiser2009care,ciocarlie2012mobile, kazhoyan2021robot,bajracharya2024demonstrating}, they generally struggle with human-robot interaction due to their reliance on predefined tasks and limited language input. They would require users to select from fixed options or explicitly re-programme the robot. 
More recent approaches leverage vision-language-based models (VLMs) to enable open-vocabulary mobile manipulation in domestic environments~\cite{brohan2023can,wu2023tidybot, yenamandrahomerobot,liu2024ok}, but they rely solely on language instructions and lack closed-loop human-robot interaction.
Another area of research explores treating robot assistants as ``physical avatars'', which allows remote users to teleoperate the robots using VR controllers\cite{de2021leveraging}, haptic devices\cite{wyrobek2008towards}, haptic gloves\cite{dafarra2024icub3}, and hand tracking\cite{chengopen}. However, these approaches can result in a high cognitive workload~\cite{hetrick2020comparing}, making them impractical for everyday use.
In this paper, we present a human-robot interaction system for remote user to naturally instruct open-vocabulary mobile manipulation with multi-round interaction using both language and gestures.

\section{Overview}
\label{sec:overview}

This work addresses the problem of remote human-robot interaction for household robot assistants. We present a multimodal system, \RobiButler{}, that combines speech commands and gesture inputs, allowing remote users to naturally guide a robotic assistant to perform household tasks.

\subsection{System Overview}
\label{subsec:overview}
The developed \RobiButler{} system is illustrated in~\figref{fig:framework}. 
It enables seamless interaction between a user wearing a Mixed Reality (MR) Head-Mounted Display (HMD) and a robot. Users can send text/voice instructions $L$ and gesture selections $G$ to the robot while receiving video streams and text/voice feedback $F$ in return.
The robotic system comprises three key components. The communication interfaces $C$ facilitate bidirectional communication, receiving user inputs and transmitting robot feedback. The high-level behavior module $H$, interprets user instructions $L$ and gesture selections ($G$) to understand the intent, generating an action sequence $P = \{a_0, a_1, ..., a_N \}$ for the robot to execute, along with a response $R$ to the user. This response can be low-level execution feedback or general information. The primitive skills $A$, provide core functionality that allows the robot to perceive and interact with the environment. These include basic mobile manipulation and Visual Question Answering (VQA) capabilities: \textit{move(), pick(), placeon(), open(), close()} and \textit{vqa()}. Note that all skills except \textit{open()} and \textit{close()} support both text and pointing queries.
\subsection{Hardware Setup}

Our system integrates multiple hardware components to facilitate effective human-robot interaction. The primary user interface is an Oculus Quest 3 MR headset, while the robotic platform consists of a Fetch mobile manipulator \cite{wise2016fetch} with a differential-drive base and a 7-dof arm. Tasks that require heavy computation are distributed between a local workstation powered by an NVIDIA RTX 4090 GPU and a remote cloud server. To enhance user visual feedback, we incorporate two additional cameras that provide third-person views of the robot's operational environment.

\section{System Implementation}
\label{sec:methods}
The system has a multimodal communication interface, a high-level behavior module, and low-level action modules.

\subsection{Communication Interfaces}
\begin{figure}[ptb]
  \centering
  \includegraphics[width=0.9\linewidth]{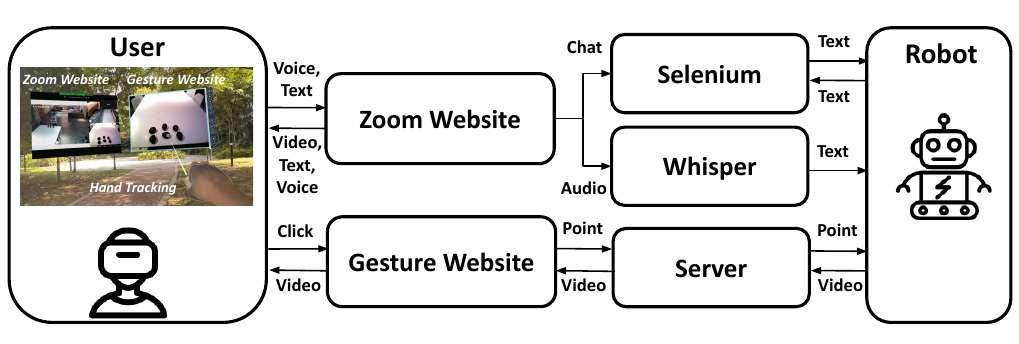}
  \caption{The framework of communication interfaces.}
  \label{fig:communicationinterface}
  \vspace{-0.6cm}
\end{figure}

As shown in~\figref{fig:communicationinterface}, the communication interfaces enable multimodal remote interaction between humans and robots, utilizing voice, text, and gestures. These interfaces consist of two main components: a Zoom platform and a gesture selection website. 
The Zoom platform supports voice, text, and video communication, while the Selenium library on the robot's server extracts specific text elements from the chat box during live sessions. For speech recognition, we employ the Whisper model \cite{radford2023robust}. 
For gesture-based interactions, we developed a website using Flask that allows users to select target objects by pointing. The site streams the robot's first-person video frames at 5 Hz, and the selected points are transmitted to the robot server in real-time, enabling immediate planning and execution. Since our design doesn't rely on a high-frequency control loop like in teleoperation, this interface is not sensitive to latency, allowing users to instruct the robot from anywhere in the world. %

\subsection{High-level Behavior Module}

\begin{figure}[hpb]
  \centering
  \includegraphics[width=\linewidth]{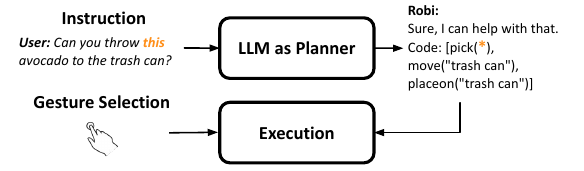}
  \caption{The framework of high-level behavior module.}
  \label{fig:highlevelframework}
  \vspace{-0.3cm}
\end{figure}

The high-level behavior module interprets and decomposes user multimodal instructions, comprising language inputs ($L$) and gesture inputs ($G$), into executable action sequences $P = \{a_0, a_1, ..., a_N | a_i \in A\}$, along with corresponding responses ($R$). This module processes both inputs, leveraging an LLM to generate structured responses and action plans. These are then passed to the execution module, which integrates the gesture inputs to ensure precise alignment between user gestures and robot actions, as depicted in~\figref{fig:highlevelframework}.

\begin{figure}[t]
\centering
\begin{subfigure}[b]{0.8\linewidth}
    \includegraphics[width=\textwidth]{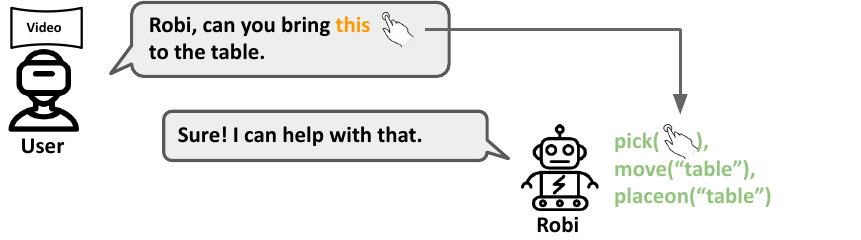}
    \caption{The user directly gives multimodal instructions to the robot.}
    \label{fig:highlevel_sub1}
\end{subfigure}

\begin{subfigure}[t]{0.9\linewidth}
\includegraphics[width=\textwidth]{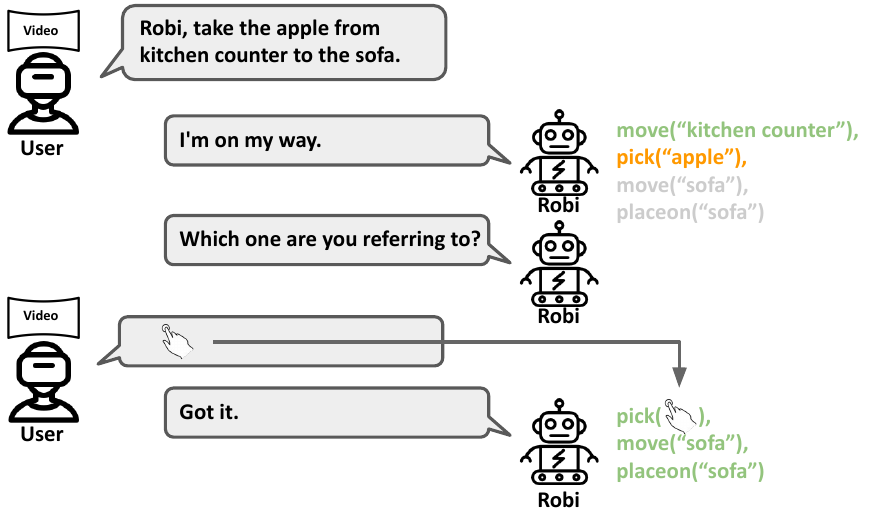}
    \caption{The user gives language and gesture instructions separately.}
    \label{fig:highlevel_sub2}
\end{subfigure}

\caption{Human-Robot Remote interactions via language and gestures. }
\label{fig:highlevel}
\vspace{-0.8cm}
\end{figure}

The task planner in the high-level behavior module, illustrated in~\figref{fig:highlevelframework}, is powered by an LLM (OpenAI GPT-4o-2024-05-13) prompted to function as a household robot assistant. The prompt defines the robot's role, a list of known locations, primitive skills it can perform, and few-shot examples to demonstrate how these skills should be used. Full prompts for the LLM can be found at \url{https://robibutler.github.io}. 
To align instructions with gesture selections, we implement a rule: when inputs contain the keywords \textit{``this''} or \textit{``here''}, the planner generates \textit{``*''} as an action parameter to resolve ambiguities, particularly demonstrative pronouns~\cite{diessel2020demonstratives}. For example, the instruction \quote{Robi, please pick this and put it on the plate} results in the plan \textit{[pick(*), placeon(``plate'')]}. During execution, the \textit{``*''} is resolved using the latest gesture selection.
We store the five most recent gesture selections and match them with the \textit{``*''} parameters during execution. Additionally, the system supports gesture-only input for disambiguation when the detection model identifies multiple objects in response to a single query. In such cases, the robot prompts, \quote{Which one are you referring to?}, pausing for the user to select the target object. Fig.~\ref{fig:highlevel} illustrates the alignment between gesture selections and the LLM-generated plan.

\subsection{Primitive Skills}

\subsubsection{Manipulation}
\label{subsubsection:manipulation}
For the robot to physically interact with the environment, it is equipped with manipulation skills such as picking/placing items, and opening/closing appliances.

\begin{figure}[htpb]
  \centering
  \includegraphics[width=0.9\linewidth]{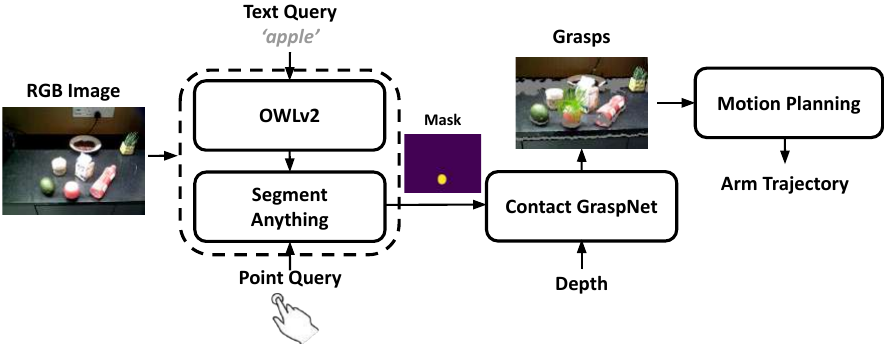}
  \caption{The open-vocabulary pick pipeline.}
  \label{fig:graspingpipeline}
  \vspace{-0.4cm}
\end{figure}

\noindent\textbf{Pick and Place Policy}
\figref{fig:graspingpipeline} illustrates the modular framework for the pick policy. The \textit{pick()} function accepts either a text query \textit{pick(text)} or a pointing query \textit{pick(point)}. We employ the pre-trained open-vocabulary detection model OWLv2~\cite{minderer2024scaling} and the Segment Anything model~\cite{kirillov2023segment} to generate the target object mask. This mask is then combined with the pre-trained grasping model Contact-GraspNet~\cite{sundermeyer2021contact} to determine grasping poses. Grasping poses are filtered based on orientation and ranked by the score.

\label{subsubsection:pick}

Given the highest-scoring grasp, a straightforward pre-grasp and grasp strategy is applied, with arm trajectories generated using the motion planning tools from MoveIt~\cite{chitta2012moveit}.

The place policy, similar to the pick policy, utilizes the same perception modules and can handle both text and pointing queries. After obtaining the segmented point clouds, the center of the place position is calculated in the X-Y plane, while the height is determined by adding 0.2 meters to the highest point of the segmented point clouds. For larger fixed objects or locations, such as tables, counters, and trash cans, a pre-defined location is used to simplify the setting.

\noindent\textbf{Open and Close Policy}
Similar to Chen et al. ~\cite{chen2023llm}, the open/close policies rely on imitation learning to handle complex actions such as opening a refrigerator and a cabinet. We collected an average of 50 demonstrations per action using teleoperation. These demonstrations were used to train the policies using Action Chunking with Transformers (ACT)~\cite{zhao2023learning}. Demonstrations of the learned skills can be viewed at \url{https://youtu.be/ajfPVjjlBcI}. \\
\subsubsection{Navigation}
\label{subsubsection:navigation}
\begin{figure}[ptb]
  \centering
  \includegraphics[width=0.97\linewidth]{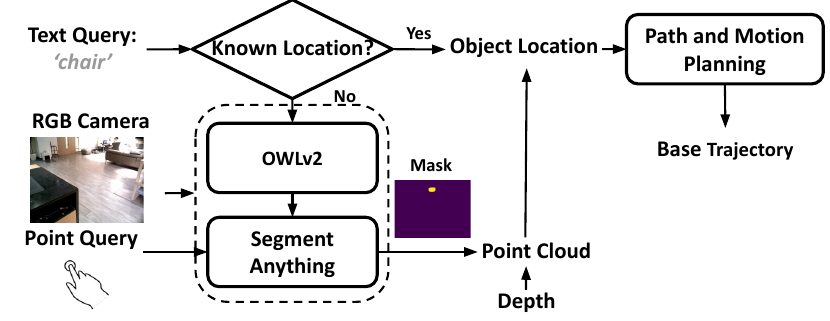}
  \caption{The navigation pipeline.}
  \label{fig:navigationpipeline}
  \vspace{-0.6cm}
\end{figure}

As shown in Fig.~\ref{fig:navigationpipeline}, our system integrates both predefined navigation places and open-world navigation to locate and move to the target object. First, we create an occupancy map using Gmapping \cite{grisetti2007improved} and define the navigation waypoint for the known locations in the map manually. 
In addition to predefined locations, the system also supports navigating to non-predefined locations via voice/text and gesture/point queries, similar to the perception pipeline in the pick policy (Sec~\ref{subsubsection:manipulation}). 
We utilize the off-the-shelf path and motion planning algorithm provided by the ROS Navigation Stack to generate the path and motion trajectory.

\subsubsection{Visual Question Answering}
\label{subsubsection:vqa}

Our system can also answer the user's open-ended questions about the status of the environment. Specifically, combined the actions \textit{vqa()}, our system supports:

\noindent\textbf{Question answering via mobile manipulation.} 
To answer the question \quote{Do we have any beer left in the fridge?}, the robot should first navigate to the fridge, open it, and then query the VLM model.
Our solution treats the VQA as a single action and uses the reasoning capabilities of LLMs to determine the necessary high-level steps before performing VQA. 
Given the question, the high-level behavior module decomposes the question into a series of actions to be executed before querying GPT-4o for the final answer.\\

\begin{figure}[ptb]
  \centering
  \includegraphics[width=0.95\linewidth]{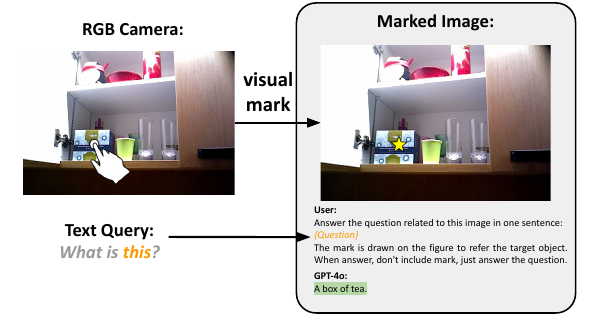}
  \caption{Example of the question answering via point referring.}
  \label{fig:vqa_pointing}
  \vspace{-0.7cm}
\end{figure}

\noindent\textbf{Question answering via point referring.}
Text-only input may lack precision, so we enable the robot to answer verbal/textual questions combined with pointing selection (\textit{vqa(text, pointing)}), as shown in Fig.~\ref{fig:vqa_pointing}.
We use a simple visual prompting method for GPT-4o to answer specific questions by annotating the image with a mark.

\begin{figure*}[tb]
  \centering
  \begin{tabular}{@{}ccccc@{}}
    \multicolumn{5}{l}{ T1: \textit{Throw \textbf{avocado} into the trash can.}}
    \vspace{0.05cm}
    \\
    \includegraphics[width = 0.19\textwidth]{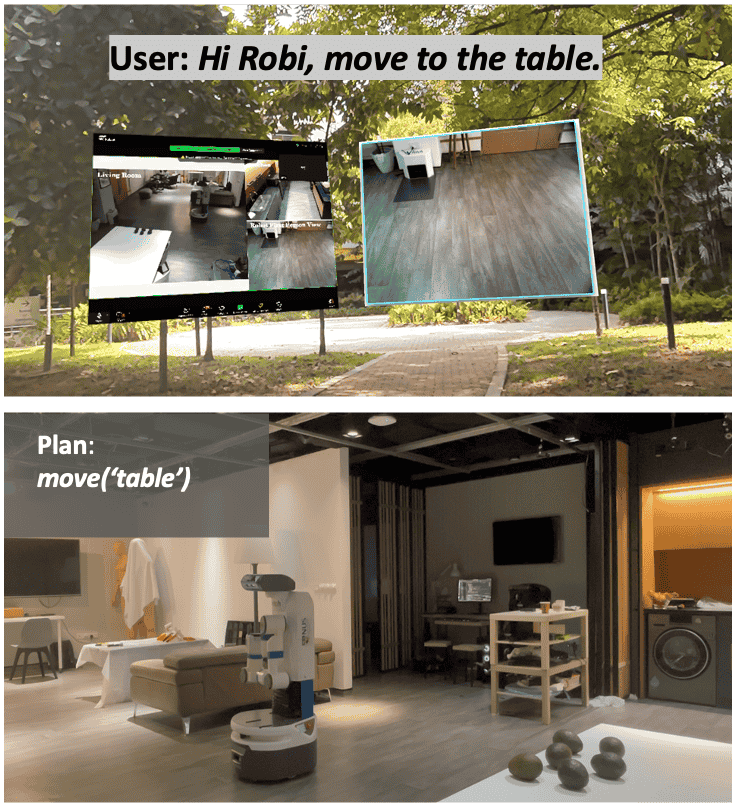} \hspace{-0.35cm}
    &
    \includegraphics[width = 0.19\textwidth]{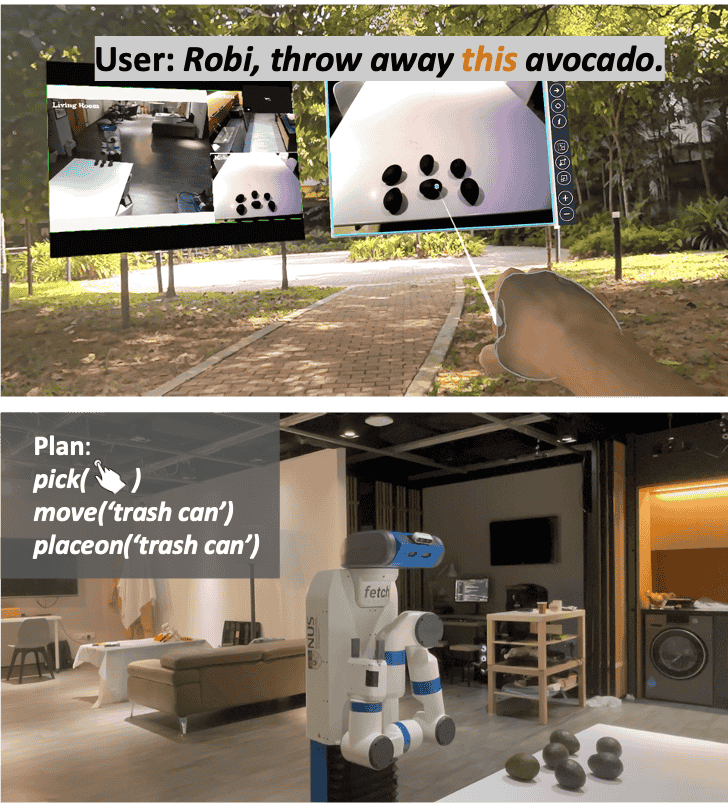} \hspace{-0.35cm}
    &
    \includegraphics[width = 0.19\textwidth]{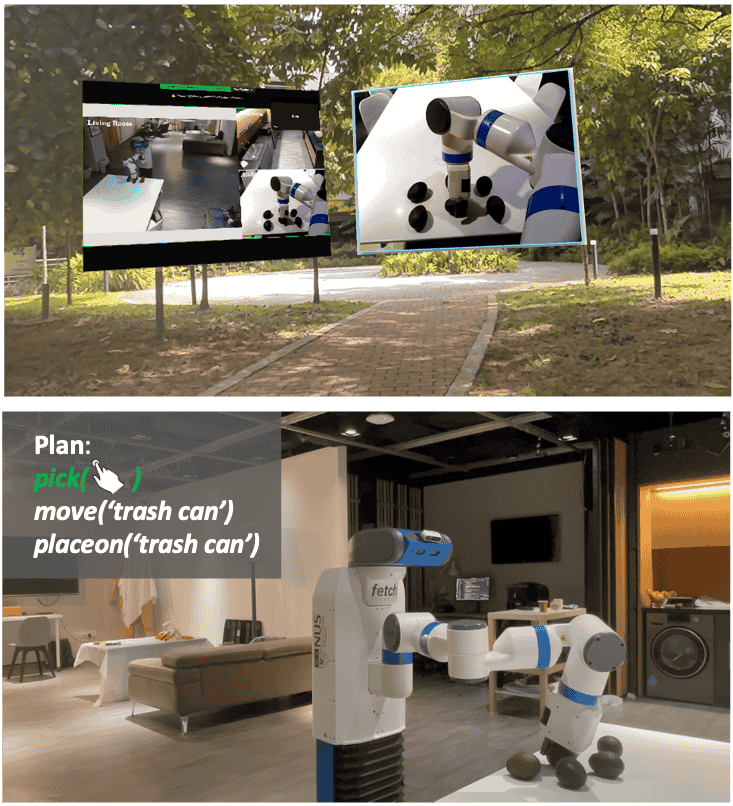} \hspace{-0.35cm}
    &
    \includegraphics[width = 0.19\textwidth]{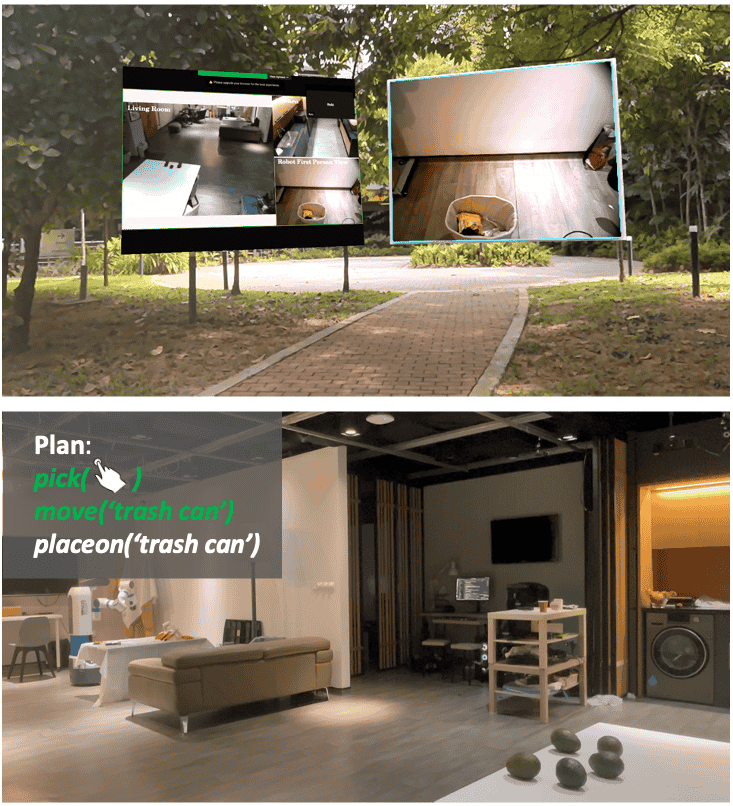} \hspace{-0.35cm}
    &    \includegraphics[width = 0.19\textwidth]{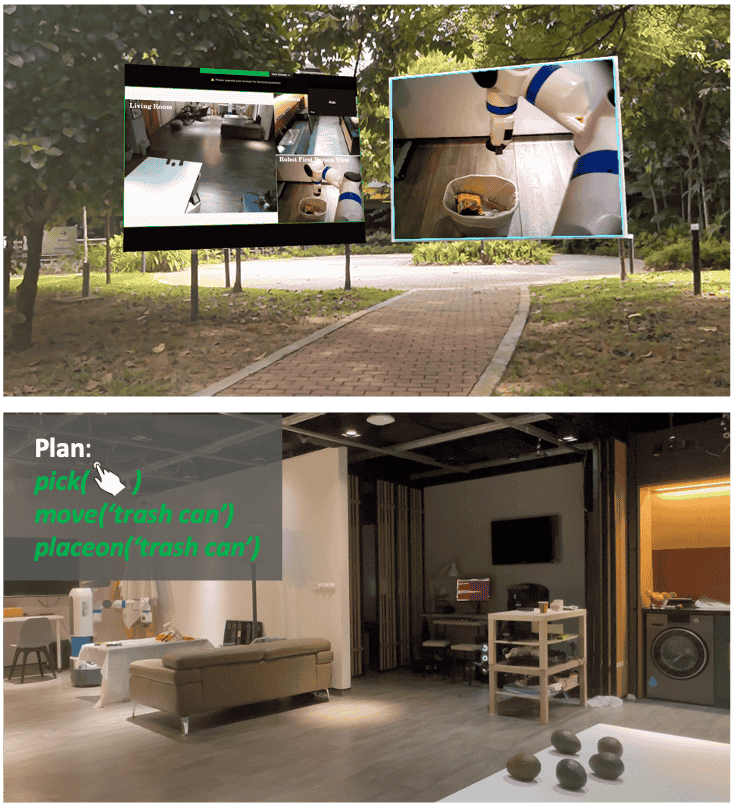} 
    \\    
    \subfig{(a.1)}
    \hspace{-0.35cm}
    & 
    \subfig{(a.2)}
    \hspace{-0.35cm}
        & 
    \subfig{(a.3)}
    \hspace{-0.35cm}
        & 
    \subfig{(a.4)}
    \hspace{-0.35cm}
        & 
    \subfig{(a.5)}
    \\ 
    \vspace{-0.4cm}
    \\
    \multicolumn{5}{l}{ T4: \textit{Describe the \textbf{object} in the cabinet.}}
        \vspace{0.05cm}

    \\
    
        \includegraphics[width = 0.19\textwidth]{figures/p11.png} \hspace{-0.35cm}
    &
    \includegraphics[width = 0.19\textwidth]{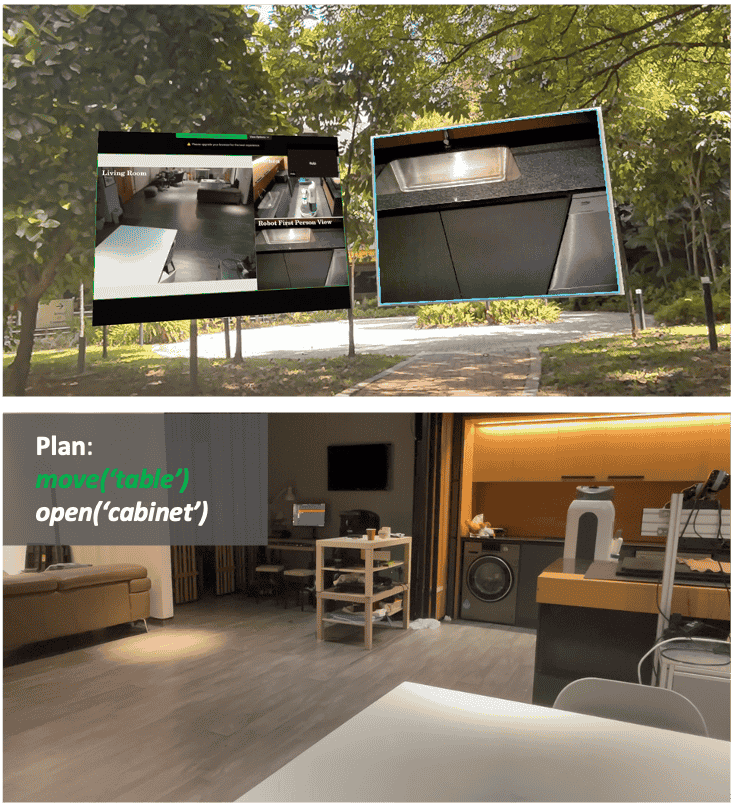} \hspace{-0.35cm}
    &
    \includegraphics[width = 0.19\textwidth]{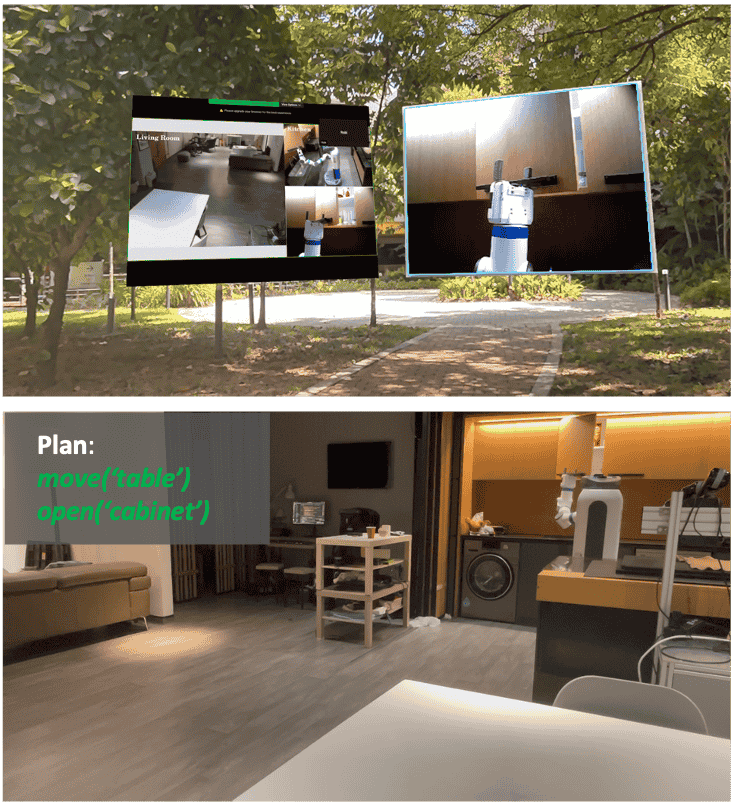} \hspace{-0.35cm}
    &
    \includegraphics[width = 0.19\textwidth]{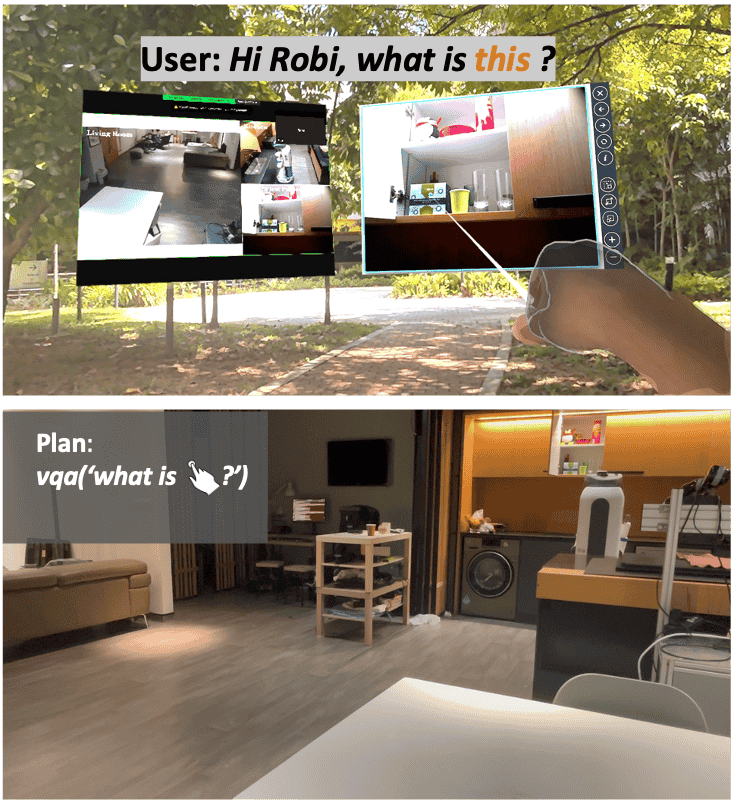} \hspace{-0.35cm}
    &    \includegraphics[width = 0.19\textwidth]{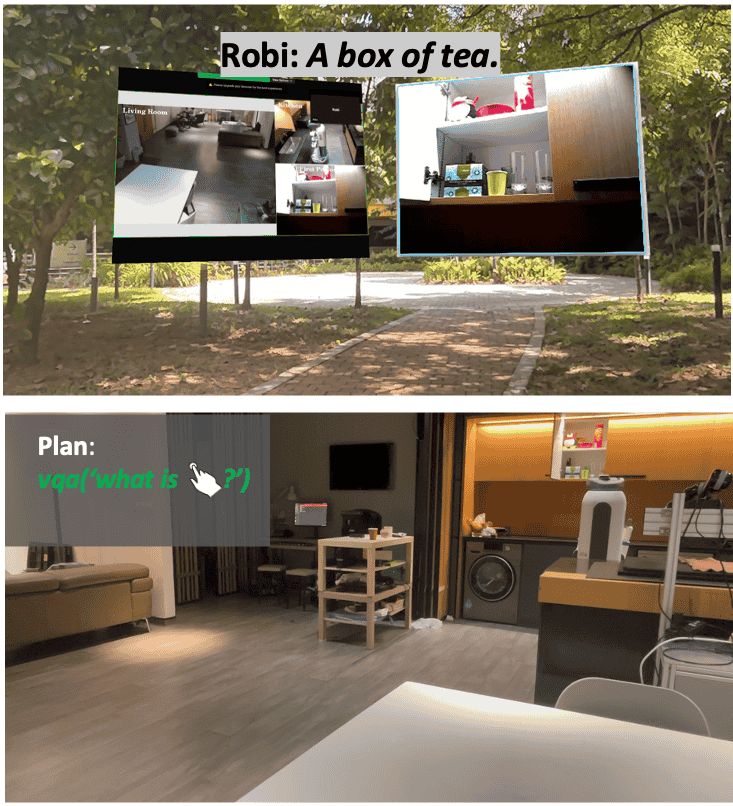} 
    \\
    \subfig{(b.1)} \hspace{-0.35cm}
    & 
    \subfig{(b.2)} \hspace{-0.35cm}
        & 
    \subfig{(b.3)} \hspace{-0.35cm}
        & 
    \subfig{(b.4)} \hspace{-0.35cm}
        & 
    \subfig{(b.5)}
    
    \vspace{-0.1cm}

  \end{tabular}
  \caption{Snapshots of completing  tasks T1 and T4. 
  \subfig{(a.1)}: User asks Robi to go to the table.
\subfig{(a.2)}: User asks Robi to throw away the avocado.
\subfig{(a.3)}: Robi picks up the avocado.
\subfig{(a.4)}: Robi brings the avocado to the trash can.
\subfig{(a.5)}: Robi throws away the avocado.
\subfig{(b.1)}: User asks Robi to open the cabinet.
\subfig{(b.2)}: Robi reaches the cabinet.
\subfig{(b.3)}: Robi opens the cabinet.
\subfig{(b.4)}: User asks Robi to identify an object.
\subfig{(b.5)}: Robi identifies it as ``A box of tea."
}
\vspace{-0.6cm}
\label{fig:experiment-snapshots}
\end{figure*}
\section{Experiments and Results}

To understand the usage and impact of multimodal remote interaction in remote HRI, we evaluate the performance of the \RobiButler{} guided by the following research questions:

\RQI{}: \textit{How effectively and robustly does the \RobiButler{} enable remote users to complete household tasks?}

\RQII{}: \textit{How do the user interaction modalities (voice, gestures) affect the performance and usability of \RobiButler{}?}

\subsection{Experiment I: \RobiButler{} Performance Evaluation}
\label{sec:experiment1}
In this experiment, we evaluate the \RobiButler{} system on a set of daily household tasks to understand its effectiveness and answer \RQI{}.
\subsubsection{Experimental Design}
\label{subsubsec:e1setting}

The tasks were designed based on the American Time Use Survey \cite{bls2019atus}. These tasks fall under the common daily household activities, including \textit{Food and drink preparation} (0.50 hr/day), \textit{Interior cleaning} (0.35 hr/day), \textit{Household \& personal organization and planning} (0.11 hr/day), and \textit{Medical and care services} (0.06 hr/day). The ten selected tasks (\textbf{T1-T10}) required the robot to interpret remote users' language and pointing gestures, then perform the corresponding actions (e.g., rearranging objects, answering questions). The object that requires disambiguation is highlighted in \textbf{bold}. To focus on remote human-robot interaction, we use objects compatible with the hardware, excluding those that are hard to manipulate, e.g., deformable or transparent.
\begin{small}
\begin{enumerate}[noitemsep, label=\textbf{T\arabic*.}, align=left, labelwidth=0.5cm]

    \item \textit{Throw \textbf{avocado} into the trash can.}
    \item \textit{Check the beer inside the fridge.}
    \item \textit{Check medicine on the coffee table and bring \textbf{one} to the sofa.}
    \item \textit{Describe the \textbf{object} in the cabinet.}
    \item \textit{Bring the \textbf{drink} to the coffee table.}
    \item \textit{Move the \textbf{cup} to the kitchen counter.}
    \item \textit{Fetch the \textbf{remote} and place it on the sofa.}
    \item \textit{Navigate to a \textbf{chair} and check if it's clean.}
    \item \textit{Check if the laptop is open.}
    \item \textit{Bring the \textbf{tool} to the table.}
\end{enumerate}
\end{small}

To evaluate the effectiveness of \RobiButler{}, the following metrics were used: \textit{\textbf{Task Success Rate (Task SR)}}: defined as the percentage of tasks completed.
\textit{\textbf{Planning Success Rate (Planning SR)}}: defined as the percentage of tasks completed when execution errors are ignored.
\textit{\textbf{Task Completion Time}}, measuring the average time required to complete each task.
\textit{\textbf{Average Interactions}}: calculating the average number of voice and gesture interactions required per task. 
A task is considered successful/completed if the goal is achieved or if correct answers are provided to the remote user within 5 minutes. 
After obtaining informed consent, the expert user evaluated \RobiButler{} on 10 tasks using free language in a fixed order, each repeated three times.

\begin{table}
\caption{Real-world Experiments Result for Experiment I. Tasks that require the user's selection are indicated using \significantI{}. Interactions include both Voice (V) and Gestures (G).}
\label{table:experiment1:results}
\centering
\resizebox{0.95\linewidth}{!}{%
\begin{tabular}{lccrc}
\toprule
\textbf{Task} & \textbf{Task SR} & \textbf{Planning SR} & \textbf{Time} & \textbf{Interactions (V + G)} \\
\midrule
\textit{T1}\significantI{} & 3/3 & 3/3  & 119.7s  & 3 (2+1)  \\
\textit{T2} & 3/3 & 3/3  & 153.0s  & 1 (1+0) \\
\textit{T3}\significantI{} & 3/3 & 3/3  & 128.3s  & 3 (1+0)  \\
\textit{T4}\significantI{} & 2/3 & 3/3  & 147.0s  & 3 (2+1) \\
\textit{T5}\significantI{} & 3/3 & 3/3  & 86.0s  & 2  (2+1)\\
\textit{T6}\significantI{} & 3/3 & 3/3  & 95.3s  & 2  (1+1)\\
\textit{T7}\significantI{} & 3/3 & 3/3  & 117.0s  & 3  (2+1)\\
\textit{T8}\significantI{} & 3/3 & 3/3  & 64.0s  & 2  (1+1)\\
\textit{T9} & 3/3 & 3/3  & 57.3s  & 2 (2+0) \\
\textit{T10}\significantI{} & 3/3 & 3/3  & 82.3s  & 2 (1+1) \\
                              \midrule
\textbf{Mean} & 96.7\% & 100\%  & 105.0s  & 2.3 (1.5 + 0.8)  \\
\bottomrule
\end{tabular}
}
\vspace{-0.7cm}
\end{table}

\subsubsection{Analysis and Results}
Table~\ref{table:experiment1:results} presents the task performance results. Overall, \RobiButler{} achieved a high average task success rate of 96.7\%, reflecting its strong ability to perform a variety of household tasks in real-world environments. However, the task success rate lags slightly behind the perfect planning success rate of 100\%, indicating challenges related to low-level action execution rather than planning processes. For instance, in task T4, an error occurred when the system misidentified a green tea box as a tissue bag.
On average, the system completed tasks in approximately 105 seconds, demonstrating its efficiency in performing household tasks in a complex environment. 
The system required an average of 2.3 interactions per task, with 1.5 voice commands and 0.8 gesture inputs. This low number of interactions demonstrates the system's efficiency in human-robot communication, requiring minimal user input to effectively guide the robot.
While the overall performance of the system is generally satisfactory, answering \RQI{}, further improvements in low-level action execution could help increase the overall performance and efficiency. Fig. \ref{fig:experiment-snapshots} shows the process of two example tasks.
The system has up to 0.2s delay in Singapore and stays within 0.5s for long distances like Singapore to Abu Dhabi.
More videos of the tasks are available at \url{https://robibutler.github.io}.

\begin{figure*}[t]
    \centering
    \includegraphics[width=1\linewidth]{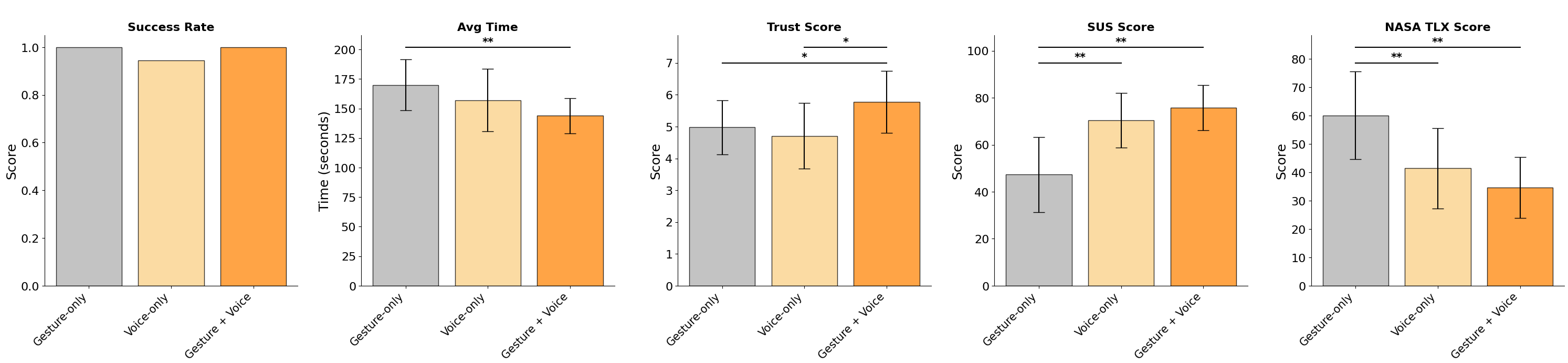}
    \caption{Measures related to efficiency and user experiences of different systems with 12 participants. For Success Rate, Trust, and SUS, the higher, the better; for Avg Time and NASA TLX, the lower, the better. For statistical significance, one asterisk (*) is \pval{<0.05}; two asterisks (**) is \pval{<0.01}.
    }
    \label{fig:combined_system_comparison}
    \vspace{-0.4cm}
\end{figure*}

\begin{figure}[t]
  \centering
  \includegraphics[width=0.98\linewidth]{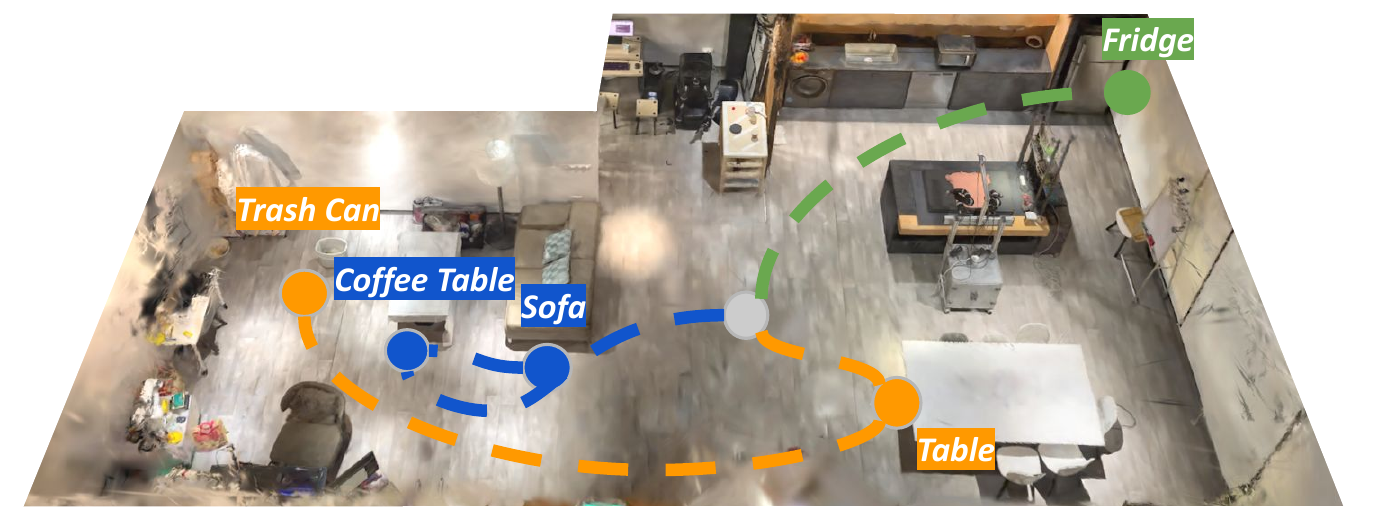}
    \caption{Visualization of the experimental environment: Orange, green, and blue trajectories represent \textit{T1}, \textit{T2}, and \textit{T3}, respectively.}
  \label{fig:rls}
  \vspace{-0.4cm}

\end{figure}

\subsection{Experiment II: The Effect of Modality on User Experience}
\label{sec:experiment2}
To investigate user experience, the impact of multimodal communication, and challenges, we conducted this experiment with novice users to address \RQII{}. 

\subsubsection{Experimental Design}

We recruited twelve volunteers (P1–P12; 7 males, 5 females) from the university community with IRB approval. None had prior experience with AR/MR smart glasses.
We compared the performance of \RobiButler{} with two baseline systems by removing user interaction modalities, similar to an ablation study, resulting in three systems: \GestureOnly{}, \VoiceOnly{}, and \RobiButler{} (\GesturePlusVoice{}). In the \GestureOnly{} system, buttons were added for participants to select the action to be executed. For the \VoiceOnly{} system, we adapted the interactive visual grounding model from \cite{xu2024TIO}. Videos of two baseline systems are available on the website.
Three representative tasks, \textit{T1} (object rearrangement), \textit{T2} (monitoring), and \textit{T3} (object rearrangement + monitoring), were selected from the previous experiment (Sec~\ref{subsubsec:e1setting}). As shown in Fig.~\ref{fig:rls}, these tasks engaged the main areas of the home. Participants were instructed to use their preferred verbal expressions.

The study used a within-subject design with three system conditions as the independent variable, counterbalanced via a Latin Square, to minimize ordering effects. Tasks increased in difficulty and were presented in a fixed order. Participants completed all three tasks with each system (nine tasks total) and filled out a questionnaire after each system to assess their subjective experience.
In addition to the \textit{Task SR} and \textit{Task Completion Time} measure from \ref{subsubsec:e1setting}, the following additional measures were used to assess user experience: 
\textit{\textbf{NASA-TLX}} \cite{nasa_tlx_2006}, assessing the perceived workload experienced by participants with each system. 
\textit{\textbf{System Usability Scale (SUS)}} \cite{brooke_sus_1996}, evaluating perceived system usability. 
\textit{\textbf{Trust}} \cite{ullman2019mdmt}, measuring the participants' trust. We used the reliable subscale under Capacity Trust.

\subsubsection{Analysis and Results}

Fig.~\ref{fig:combined_system_comparison} shows the task performance of the three systems. 
A one-way repeated measures ANOVA was conducted to analyze the quantitative data after confirming normality assumptions.
Both the \GestureOnly{} and \GesturePlusVoice{} (i.e., \RobiButler{}) systems achieved a perfect task success rate of 100\%, while the voice system had a slightly lower, though non-significant, success rate of 94.4\%. This difference was attributed to errors in target referencing with voice commands only. For example, the voice recognition system misinterpreted the word `right' as `red',  leading to the grounding error.
Additionally, \RobiButler{} ($M = 143.8$, $SD=14.8$) had a significantly lower task completion time than the \GestureOnly{} system ($M=170.00$, $SD=21.4$) (\pval{<0.05}), but was not significantly lower than the \VoiceOnly{} system ($M=157.1$, $SD=26.6$). The reduced task completion time for voice-supported systems primarily resulted from the ability to use voice commands to express combined queries, whereas with the \GestureOnly{} system, participants had to perform multiple manual clicks, increasing task completion time.

Regarding the trust, the \RobiButler{} ($M = 5.77$, $SD = 0.97$) was perceived as significantly more trustworthy compared to both the \GestureOnly{} system ($M = 4.98$, $SD = 0.85$, \pval{<0.05}) and the \VoiceOnly{} system ($M = 4.71$, $SD = 1.03$, \pval{<0.05}). 
This suggests that combining gestures with voice enhances the system's reliability and consistency, outperforming single-modality systems.
P2 reasoned that \quote{I trusted the gesture plus voice system the most because I found it easier to avoid making mistakes with it. For language only, sometimes it may misunderstand me. For gestures, I have to do the interaction multiple times.}
For the SUS, participants gave the \GestureOnly{} the lowest usability score ($M = 47.29$, $SD = 15.90$), which significantly lower than both \VoiceOnly{} ($M = 70.42$, $SD = 11.62$, \pval{<0.01}) and \RobiButler{} ($M = 75.83$, $SD = 9.61$, \pval{<0.01}). This also indicates that \RobiButler{} achieved `Good' usability (i.e., $SUS > 75$ ~\cite{bangor_empirical_2008}) compared to the other systems. 

Overall, the \RobiButler{} achieves the best performance, the highest usability, and the minimum perceived cognitive load among the baselines, answering \RQII{}. 
This was primarily due to the complementary nature of voice and gestures—voice enabled natural queries, while gestures provided precise spatial annotations.
Although multimodal interaction generally outperformed unimodal interaction, P10 expressed a negative sentiment, stating, \quote{Using both voice and gesture is [sometimes] hard, as I need to switch between two modalities. I prefer voice-only as I don't need to move my arm physically.} 
Incorporating eye gaze tracking could reduce physical workload by minimizing hand interactions.

\section{Conclusion}
\label{sec:conclusion}

This work introduces an interactive robotic assistant for household tasks using multimodal interactions with remote users. We outline three core components of the robot butler system and demonstrate its effectiveness in assistive question-answering and object rearrangement tasks. Experiments show \RobiButler{} grounds remote multimodal instructions with a high task success rate, reasonable time, and minimal interactions. Follow-up tests confirm that combining voice and gestures enhances usability and trust, and reduces cognitive load compared to unimodal systems. 
In future work, we aim to make \RobiButler{} more adaptable, capable of lifelong learning, personalized interactions, and handling complex tasks that may require tactile feedback \cite{yu2024octopi}.

\section*{Acknowledgment}
We would like to thank Yuhong Deng, Hanbo Zhang, and Tongmiao Xu for their invaluable assistance in model development and data collection. We also sincerely appreciate all the volunteers who gave their time to participate in the human-robot interaction experiments.
This research is supported in part by the National Research Foundation (NRF), Singapore and DSO National Laboratories under the AI Singapore Program (AISG Award No: AISG2-RP-2020-016) and by the Agency for Science, Technology \& Research (A*STAR), Singapore  under its National Robotics Program (No. M23NBK0091).

{
    \bibliographystyle{IEEEtran}
    \bibliography{IEEEabrv, bib/bibliography.bib}
}

\end{document}